\documentclass[sigconf]{acmart}
\AtBeginDocument{%
  }

\copyrightyear{2026}
\acmYear{2026}
\setcopyright{cc}
\setcctype{by}
\acmConference[MM '26]{Proceedings of the 34th ACM International Conference on Multimedia}{November 10--14, 2026}{Rio de Janeiro, Brazil}
\acmBooktitle{Proceedings of the 34th ACM International Conference on Multimedia (MM '26), November 10--14, 2026, Rio de Janeiro, Brazil}
\acmDOI{10.1145/3767308.3835803}
\acmISBN{979-8-4007-2213-4/2026/11}

\usepackage{graphicx}
\usepackage{float} 
\usepackage{subfigure}
\usepackage{algorithm}
\usepackage[noend]{algpseudocode} % algpseudocode 宏包功能更强，推荐使用
\usepackage{threeparttable} % 表格
\usepackage{multirow} % 表格跨行单元格
\usepackage{xspace}
\usepackage{tabularx}

\usepackage{tcolorbox}  % 核心：生成带标题栏的框
\usepackage{xcolor}     % 控制背景/字体颜色
\usepackage{ragged2e}   % 内容自动换行（适配页宽）
\usepackage{makecell}

\newtcolorbox{promptbox}{
    colback=white!90!gray,        % 内容区背景：白色
    colframe=black,       % 边框颜色：黑色
    halign upper=justify, % 内容区两端对齐
    boxrule=0.8pt,        % 边框粗细
    width=\linewidth      % 宽度占满页宽
}

\usepackage{xurl} 

\begin{document}

\newcommand{\etal}{{\em et al.}\xspace}
\newcommand{\ie}{{\em i.e.},\xspace}
\newcommand{\eg}{{\em e.g.},\xspace}

\title{SwipeGen: Bridging the Execution Gap in GUI Agents via Human-like Swipe Synthesis}

  
\author{Xuan Wang}
\authornote{These authors contributed equally to this work.}
\affiliation{
  \department{College of Computer Science and Artificial Intelligence}
  \institution{Fudan University}
  \city{Shanghai}
  \country{China}}
\email{xuanwang25@m.fudan.edu.cn}

\author{Siyuan Su}
\authornotemark[1]
\affiliation{
  \department{College of Computer Science and Artificial Intelligence}
  \institution{Fudan University}
  \city{Shanghai}
  \country{China}}
\email{24300810015@m.fudan.edu.cn}

\author{Quantong Fu}
\authornotemark[1]
\affiliation{
  \department{College of Computer Science and Artificial Intelligence}
  \institution{Fudan University}
  \city{Shanghai}
  \country{China}}
\email{qtfu22@m.fudan.edu.cn}

\author{Yongxiang Hu}
\affiliation{
  \department{College of Computer Science and Artificial Intelligence}
  \institution{Fudan University}
  \city{Shanghai}
  \country{China}}
\email{huyx17@fudan.edu.cn}
  
\author{Yangfan Zhou}
\correspondingauthor
\affiliation{
  \department{College of Computer Science and Artificial Intelligence}
  \institution{Fudan University }
  \city{Shanghai}
  \country{China}}
\email{zyf@fudan.edu.cn}

\begin{abstract}
Despite numerous Graphical User Interface (GUI) agents claiming to automate user interaction tasks, to date, few achieve satisfactory interaction capability with human users in real-world scenarios.
Through empirical analysis, this paper identifies the root cause of the limited interaction capability as the rigid swipe execution.
In particular, unlike humans, who perform swipes with fine-grained control over trajectory, speed, and timing, existing agents can only conduct simplistic, deterministic swipe behaviors, leading to frequent failures on complicated user-like interaction tasks.
Due to the lack of open-source human-like swipe training data, we propose SwipeGen, the first tool for synthesizing diverse and human-like swipe interactions, and SwipeBench, the first benchmark for evaluating agents' swipe interaction quality.
Extensive experiments show that SwipeGen can improve the swipe execution success rate of existing agents by up to 2.46$\times$.
Our code, dataset, and model are available at \url{https://github.com/TSKGHS17/SwipeGen}.
\end{abstract}




\maketitle

\section{Introduction}
\label{sec:intro}

% 1. Background
Graphical User Interface (GUI) agents~\cite{chen2025guicourse, nguyen2025guiagentsurvey} are autonomous systems that interact with GUIs in a human-like manner based on natural language commands. These agents have found applications in various domains, such as mobile assistants~\cite{minh24gptvoicetasker, wang2024mobileagentv2, zhang2025appagent}, accessibility tools~\cite{ram24savant, peng2025morae}, and automated GUI testing~\cite{hu2024auitest, liu25visiondroid, feng2025tiktokagent, wang25testweaver}.
Recently, with the rise of Vision-Language Models (VLMs), VLM-based GUI agents~\cite{cheng2024seeclick, hong2024cogagent, kevin2025showui, wu2025osatlas, gou2025uground, lu2025uir1, luo2025guir1} have become increasingly popular.
This is because VLMs can process both visual (\eg GUI images) and textual (\eg natural language commands) information. 
The integration of multimodal inputs significantly enhances the decision making capability of GUI agents, enabling more reliable and consistent performance.

\begin{figure*}[t]
  \includegraphics[width=\linewidth]{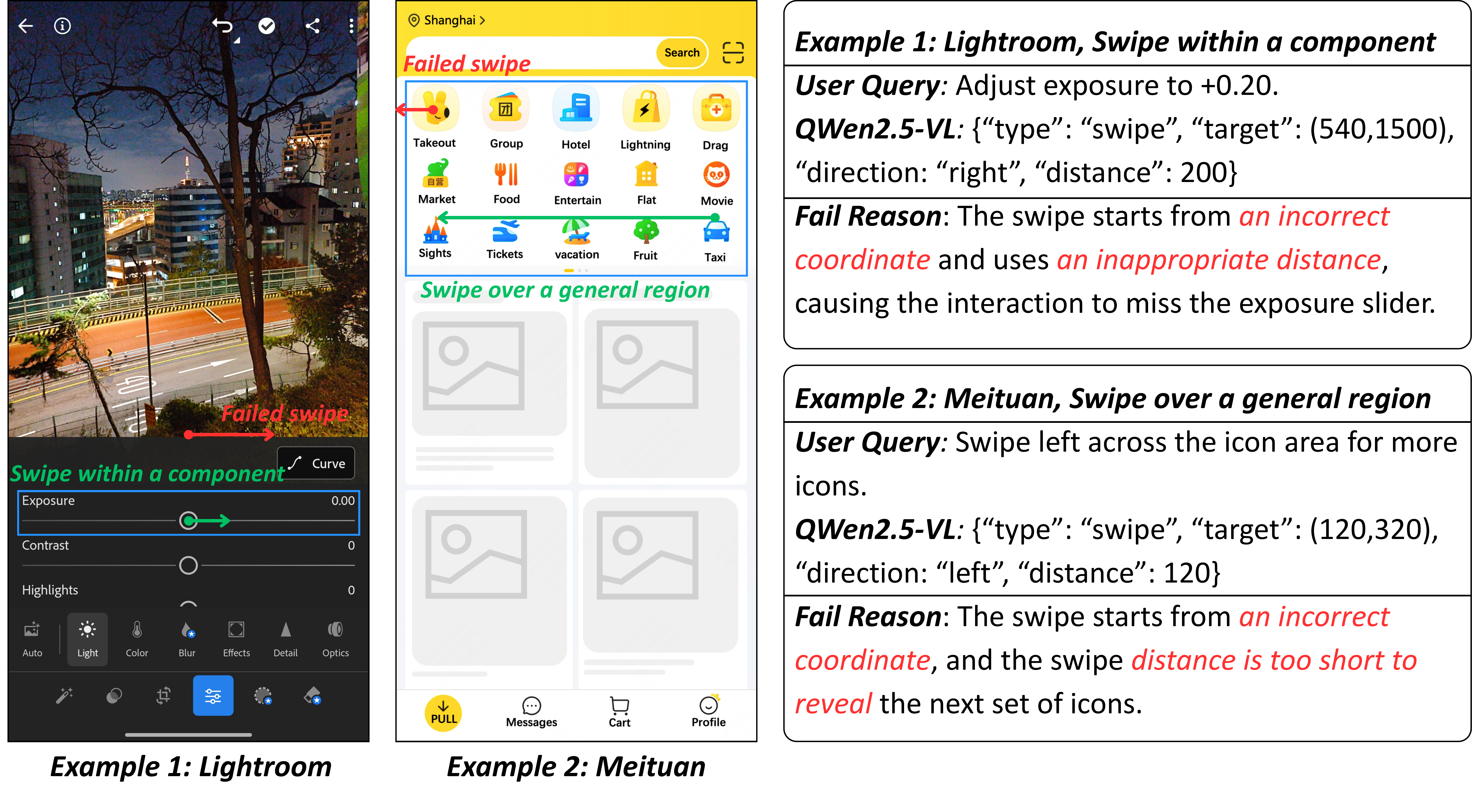}
  \vspace{-9pt}
  \caption{Two common types of swipes in mobile GUIs.}
  \label{fig:1_intro}
  \vspace{-3pt}
\end{figure*}

However, such improvements primarily stem from better GUI semantic understanding, while the generated GUI interactions still deviate substantially from human interaction patterns.
Specifically, existing GUI agents still struggle to perform \textit{swipe}, one of the most frequently used interactions.
A key reason is that most GUI agents \textit{implicitly assume} GUI interactions to be \textit{component-centric}.
They formulate GUI interaction as an action prediction task that maps a natural language command (\eg tap a button, type text into an input field) to an action type $t$ and the coordinate $(x, y)$ of a target GUI component.
Some approaches~\cite{cheng2024seeclick, gou2025uground, chen2025uiins, tang2025guig2} even simplify this formulation as a pure GUI grounding task, \ie predicting only the target component's coordinate $(x, y)$.

Such overly simplified, component-centric interaction strategy \textit{does not hold} for swipes.
As shown in Figure~\ref{fig:1_intro}, swipes in practice can be broadly categorized into two types:
1) \textit{Swiping within a component} for fine-grained adjustments (\eg dragging the exposure slider in Lightroom\footnote{\url{https://play.google.com/store/apps/details?id=com.adobe.lrmobile&hl=en}}), and 
2) \textit{Swiping over a general region} for content exploration (\eg swiping to reveal additional options in a lifestyle app, Meituan\footnote{\url{https://play.google.com/store/apps/details?id=com.sankuai.meituan&hl=en}}).
Unlike clicks and text inputs, swipes are not necessarily anchored to specific components.
As a result, existing VLM-based agents struggle to handle swipes under current task formulations, preventing them from completing a wide range of everyday tasks.

% 3. Challenge
Addressing this issue requires overcoming two fundamental obstacles:
1) \textit{Scarcity of annotated swipe data.}
Existing GUI interaction datasets~\cite{li2020androidhowto, andrea2022motif, lu2024guiodyssey, cheng2024seeclick, wu2025osatlas, chai2025amex, chen2025guicourse, zhang2025agentcpmgui, li2025screenspotpro, rawles2025androidworld} are heavily dominated by component-centric actions(\eg clicks and text inputs), which account for 76.4\% to 94.9\% of all interactions, while swipes are sparsely annotated.
2) \textit{Improper formulation of swipe.}
Unlike clicks, a swipe requires accurately predicting multiple parameters, including the starting position, ending position, direction, and duration.
However, the few datasets that do include swipe interactions~\cite{li2020androidhowto, andrea2022motif, lu2024guiodyssey, zhang2025agentcpmgui, chai2025amex} typically either ignore these parameters or oversimplify them (\eg containing only a coarse direction).
As a result, fine-tuned VLMs can hardly ground swipes correctly or generate valid swipe parameters when such interactions are required.

% 4. Method
In this paper, we present \texttt{SwipeGen}, an automatic pipeline for synthesizing human-like swipe data without relying on predefined human commands.
To properly formulate swipe interactions, \texttt{SwipeGen} decomposes each swipe into multiple execution dimensions, including the starting position, direction, distance, and velocity, consistent with widely used mobile automation tools~\cite{url:adb, url:appium, url:uiautomator}.
Based on this definition, \texttt{SwipeGen} then automatically explores GUIs to synthesize human-like swipes and record all execution-required parameters.
Specifically, it first detects scrollable targets (components and regions) on the screen, then executes candidate swipes, and finally verifies their validity by comparing GUI states before and after the interaction.
Finally, \texttt{SwipeGen} retains only swipes that induce visual changes, enabling high-quality data collection.

Based on this pipeline, we construct and release \texttt{SwipeBench}, the first benchmark for evaluating the swipe execution capability of GUI agents. 
Furthermore, we demonstrate the effectiveness of \texttt{SwipeGen} by fine-tuning an open-source VLM for GUI interaction, resulting in a swipe-capable VLM, \texttt{GUISwiper}.
\texttt{GUISwiper} achieves a swipe success rate of 61.25\% on \texttt{SwipeBench} while maintaining competitive grounding performance.
These results show that improving swipe capability does not come at the cost of degraded grounding performance, and validate the effectiveness of \texttt{SwipeGen}.

\definecolor{mygreen}{RGB}{0, 150, 0}
\definecolor{myred}{RGB}{200, 0, 0}

\begin{table*}[t]
\centering
\resizebox{\textwidth}{!}{
\begin{tabular}{lccccc ccccc}
\toprule
\multirow{2}{*}{\textbf{Dataset}} 
& \multirow{2}{*}{\textbf{\#Interactions}} 
& \multicolumn{3}{c}{\textbf{Action Distribution (\%)}} 
& \multicolumn{5}{c}{\textbf{Swipe Annotation}} \\
\cmidrule(lr){3-5} \cmidrule(lr){6-10}
& & \textbf{Click+Text} & \textbf{Swipe} & \textbf{Other} 
& \textbf{Description} & \textbf{Start Pos} & \textbf{End Pos} & \textbf{Direction} & \textbf{Duration} \\
\midrule
AndroidHowTo~\citep{li2020androidhowto}      
& 136,023 & 94.9 & 5.1  & 10.9 & \textcolor{mygreen}{$\checkmark$} & \textcolor{myred}{$\times$}  & \textcolor{myred}{$\times$} & \textcolor{myred}{$\times$} & \textcolor{myred}{$\times$} \\
MoTIF~\citep{andrea2022motif}             
& 12,244 & 94.1 & 5.9  & 0.0 & \textcolor{myred}{$\times$} & \textcolor{mygreen}{$\checkmark$} & \textcolor{mygreen}{$\checkmark$} & \textcolor{mygreen}{$\checkmark$} & \textcolor{myred}{$\times$} \\
GUI-Odyssey~\citep{lu2024guiodyssey}       
& 110,056 & 89.4 & 9.5 & 6.9 & \textcolor{myred}{$\times$} & \textcolor{mygreen}{$\checkmark$} & \textcolor{mygreen}{$\checkmark$} & \textcolor{myred}{$\times$} & \textcolor{myred}{$\times$} \\
CAGUI~\citep{zhang2025agentcpmgui}             
& 3,916 & 97.4 & 2.0 & 0.6 & \textcolor{myred}{$\times$} & \textcolor{mygreen}{$\checkmark$} & \textcolor{mygreen}{$\checkmark$} & \textcolor{myred}{$\times$} & \textcolor{myred}{$\times$} \\
AMEX~\citep{chai2025amex}              
& 35,661 & 76.4 & 21.4 & 2.2 & \textcolor{myred}{$\times$} & \textcolor{mygreen}{$\checkmark$} & \textcolor{mygreen}{$\checkmark$} & \textcolor{myred}{$\times$} & \textcolor{myred}{$\times$} \\
% \midrule
% \texttt{SwipeBench} (Ours) & TBD & 0.0 & 100.0 & 0.0 & \textcolor{mygreen}{$\checkmark$} & \textcolor{mygreen}{$\checkmark$} & \textcolor{mygreen}{$\checkmark$} & \textcolor{mygreen}{$\checkmark$} & \textcolor{mygreen}{$\checkmark$} \\
\bottomrule
\end{tabular}
}
\caption{
\textbf{Action distribution and swipe parameter annotation across representative GUI navigation datasets.}
Most existing datasets are dominated by component-centric actions (click and text input), while swipes are sparsely annotated. (\textbf{Other} denotes dataset-specific actions such as system-level back or composite gestures.)
}
\label{table:2_related}
\end{table*}

% 5. Contribution (贡献不是你做了什么，是有什么用)
Our primary contributions are as follows:
\begin{itemize} 
    \item By comparing the success rates of swipe and other interactions, we are the first to pinpoint the widely-adopted rigid swipe execution strategy as a key bottleneck limiting the interaction capability of existing GUI agents.

    \item We design \texttt{SwipeGen}, the first automated pipeline for synthesizing human-like swipe interactions. Moreover, in order to advance research on swipe-aware GUI agents, we built \texttt{SwipeBench}, the first benchmark for evaluating swipe interaction quality.
    
    \item We prove that VLMs fine-tuned with \texttt{SwipeGen}-synthesized data can substantially address the swipe execution bottleneck in GUI agents. Experiments show that \texttt{GUISwiper} improves swipe success rate by 2.46$\times$, shedding light on the development of more capable GUI agents.

\end{itemize}

\section{Related Work}
\label{sec:related}

\subsection{VLM for GUI Agents}
Despite the strong capabilities of general-purpose large VLMs like GPT-4V~\cite{openai2023gpt4v}, their performance in understanding and interacting with GUIs remains limited~\cite{yan2023gpt4vwonderland}. 
This limitation has motivated researchers to fine-tuning open-source VLMs, such as the Qwen-VL series~\cite{Qwen-VL, Qwen2VL, Qwen2.5-VL, qwen3}, to build GUI-specific models that better comprehend user commands and GUI images.

Training these models generally follows two paradigms: early-stage supervised fine-tuning (SFT) and more recent reinforcement learning (RL) approaches.
Under the SFT paradigm, models such as SeeClick~\cite{cheng2024seeclick}, CogAgent~\cite{hong2024cogagent}, OS-Atlas~\cite{wu2025osatlas}, UGround~\cite{gou2025uground}, ShowUI~\cite{kevin2025showui}, and UI-TARS~\cite{qin2025uitars} demonstrate improved GUI understanding through supervised training on large, labeled datasets.
However, SFT heavily depends on high-quality annotations, resulting in high training costs.
With the emergence of DeepSeek-R1~\cite{deepseek2025deepseekr1}, researchers found that reinforcement learning with verifiable rewards (RLVR), particularly group relative policy optimization (GRPO)~\cite{shao2024grpo}, is well suited for GUI navigation tasks.
Models such as UI-R1~\cite{lu2025uir1}, GUI-R1~\cite{luo2025guir1}, BTL-UI~\cite{zhang2025btlui} and GUI-Owl~\cite{ye2025mobileagentv3} leverage predefined, task-specific reward functions to train more capable GUI-specific VLMs.

Overall, the performance of GUI-specific VLMs is strongly tied to \textit{the distribution and quality of their training data}.
This observation motivates our work, which targets the lack of diverse and executable swipe data by proposing a scalable pipeline to enhance the swipe capabilities of VLM for GUI agents.

% \subsection{Beyond Clicking: Drag Interactions}
% \label{ssec:related_drag}

% Recent work has begun to explore more sophisticated interactions such as drags, which, akin to swipes, require fine-grained spatial control beyond single-point grounding. 
% GUI-DRAG~\cite{liao25textdrag} targets text dragging on PC through an automated data generation pipeline and comprehensive benchmark. 
% ShowUI-$\pi$~\cite{hu2025showuipi} is a flow-based generative model inspired by robotic dexterous hands. It produces continuous mouse trajectories, addressing the limitation of discrete action formulations that cannot adapt to dynamic visual feedback.

% While these advances improve drag capabilities in desktop environments, they do not fully extend to \textit{mobile swipe interactions}, which present unique challenges due to OS-level inertia dynamics~\cite{yamanaka2024swipemodel}.
% Our work bridges this gap by decomposing swipes into quantifiable dimensions and introducing the first automated pipeline and benchmark specifically designed for mobile swipe quality evaluation.

\subsection{Existing GUI Datasets}
\label{ssec:related_datasets}

\begin{figure*}[t]
  \includegraphics[width=\linewidth]{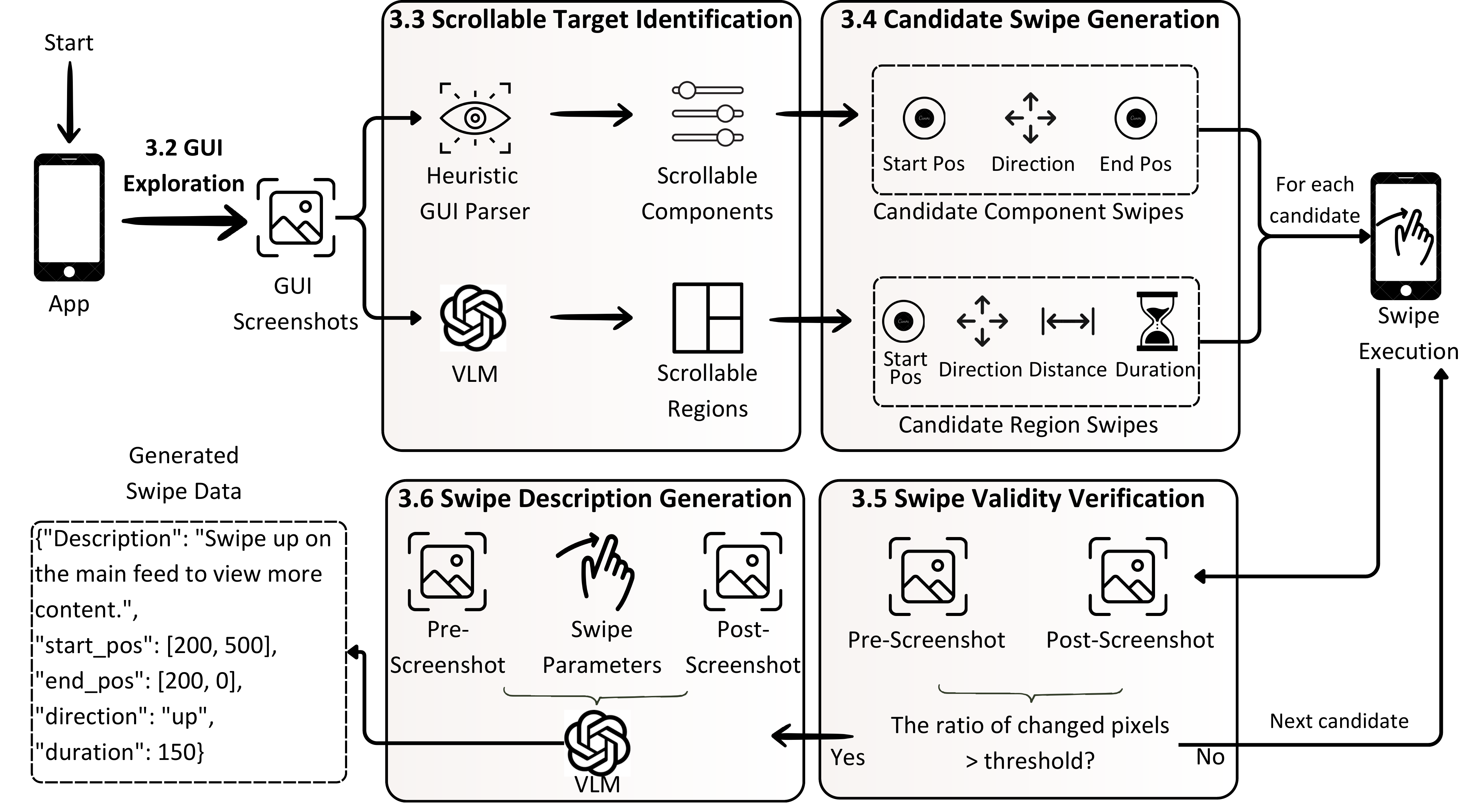}
  \caption{\textbf{Overview of the proposed pipeline \texttt{SwipeGen}.} It consists of five important modules: GUI Exploration Module, Scrollable Target Identification Module, Candidate Swipe Generation Module, Swipe Validity Verification Module and Swipe Description Generation Module.}
  \label{fig:3_swipegen}
  \vspace{-3pt}
\end{figure*}

Existing GUI datasets can be broadly categorized into GUI grounding datasets and GUI navigation datasets.
The GUI grounding task aims to locate the target GUI component given a natural language command.
Under this component-centric formulation, GUI grounding datasets naturally emphasize click and text input actions, excluding swipes.
Representative benchmarks such as ScreenSpot~\cite{cheng2024seeclick} and its subsequent variants~\cite{wu2025osatlas, li2025screenspotpro} follow this paradigm and therefore do not annotate swipes.

In contrast, GUI navigation datasets aim to model multi-step interaction trajectories for accomplishing high-level tasks.
However, as summarized in Table~\ref{table:2_related}, existing navigation datasets suffer from several limitations that hinder their use for training reliable swipes.

First, the \textit{most critical limitation} is the lack of step-level natural language supervision. Most navigation datasets provide only a single high-level task description (\eg book a hotel near the city center) paired with a sequence of annotated low-level interactions.
Individual actions, including swipes, are \textit{not associated with corresponding descriptions}, making these datasets unsuitable for training single-step swipe prediction models.
Second, the action distribution is heavily skewed toward component-centric interactions such as clicks and text inputs.
Third, even when swipes are included, their annotations are often incomplete.
As shown in Table~\ref{table:2_related}, existing datasets typically miss one or more parameters required for executable swipes, such as explicit direction or duration.
Although the swipe direction can be inferred from the start and end positions, we argue that explicitly annotating direction is important.
Moreover, swipe duration directly affects the execution outcome due to OS-level gesture dynamics~\cite{yamanaka2024swipemodel} (\eg inertia).
We would further discuss the role of these parameters in Section~\ref{ssec:unified_swipe}.

Overall, these limitations motivate the need for data that provides step-level language description and complete parameter annotations for swipes.

\section{SwipeGen}
\label{sec:approach}

This section introduces \texttt{SwipeGen}, an automatic pipeline for synthesizing executable swipe data without relying on predefined commands.
Figure~\ref{fig:3_swipegen} provides an overview of the pipeline.

Given a GUI screen, \texttt{SwipeGen} first identifies scrollable targets(\ie components and regions).
Based on the identified targets, \texttt{SwipeGen} generates and executes candidate swipes under a unified swipe representation.
To ensure data quality, each executed swipe is verified by comparing GUI images before and after it.
For each valid swipe, \texttt{SwipeGen} generates a description based on the pre- and post-swipe images together with the swipe parameters.

\subsection{Unified Swipe Representation}
\label{ssec:unified_swipe}
Before introducing \texttt{SwipeGen}, a fundamental question is unanswered: what parameters should a swipe annotation include for practical deployment?
% As summarized in Section~\ref{ssec:related_datasets}, swipes' annotations in existing GUI datasets are inconsistent and often incomplete. This makes them difficult to be directly used for training GUI-specific VLMs.

For \texttt{SwipeGen}, our guiding principle is that synthesized swipe data should be \textit{directly executable} in real-world GUI agent systems.
In GUI agent systems, the outputs of VLMs are consumed by downstream GUI automation tools to interact with GUIs~\cite{zhang2025appagent, wang2024mobileagentv2}.
Therefore, a valid swipe representation must contain sufficient parameters to invoke these tools.

To this end, we survey widely used mobile GUI automation tools for both Android and iOS, including Android Debug Bridge (ADB)~\cite{url:adb}, UIAutomator~\cite{url:uiautomator}, and Appium~\cite{url:appium}.
And we found out that all these tools parameterize swipes using explicit start and end positions, and two of them (ADB and Appium) further support specifying the swipe duration.

We then analyze how these parameters affect different types of swipe interactions.
For component-centric swipes that aim to make fine-grained adjustments (\eg adjusting a slider), the execution outcome is determined by the start and end positions.
In contrast, swipes over general regions exhibit different characteristics.
As illustrated in Figure~\ref{fig:1_intro}, the resulting scroll behavior is mainly effected by the starting position, the swipe direction, and the duration.
Notably, duration directly controls the swipe velocity, which in turn determines how far the content scrolls.
As a result, even with identical start and end positions, varied swipe duration can lead to different effect. And the exact endpoint is less important for region-level swipes. 
Therefore, although the direction can be derived from the start and end positions, we explicitly annotate it instead to facilitate learning controllable swipes.

Based on these observations, we unify swipes using four explicit parameters: start position, end position, direction, and duration.

\subsection{GUI Exploration}
To collect diverse GUI screens for subsequent swipe data generation, \texttt{SwipeGen} adopts a depth-first search (DFS)-based exploration strategy. 
Starting from the app's entry point, it systematically navigates the GUI by interacting with clickable components. 
For each visited screen, \texttt{SwipeGen} records the sequence of interactions required to reach it, enabling efficient revisiting for later scrollable target identification.

For each screen, \texttt{SwipeGen} identifies unvisited clickable components based on the underlying GUI hierarchy files (\eg XML layouts~\cite{url:UI_layout}) and executes interactions to discover new screens.
For each visited screen, \texttt{SwipeGen} records the interaction path for efficient revisiting.
% Specifically, it parses the XML file of the current screen and extracts nodes whose \texttt{clickable} attribute is set to \texttt{true}, treating them as candidates for interaction. 
% At each step, \texttt{SwipeGen} selects an unvisited clickable element and performs a click action to trigger navigation.

To distinguish newly discovered screens from previously visited ones, we represent each screen using a lightweight skeleton abstraction that reduces the influence of dynamic content.
\texttt{SwipeGen} compares the skeleton similarity between screens and continues exploration until no new screens are discovered.

% After each interaction, \texttt{SwipeGen} determines whether a new screen has been reached by comparing its structural layout with those of previously visited screens. 
% To reduce the impact of dynamic content (\eg text and images), each screen is represented using a skeleton abstraction that captures the layout of GUI components. 
% This abstraction is constructed via edge detection and morphological operations to produce a coherent structural representation. 
% \texttt{SwipeGen} then computes the similarity between skeletons to identify newly discovered screens and continues exploration accordingly.

\subsection{Scrollable Target Identification}
For each collected GUI screen, \texttt{SwipeGen} identifies two types of scrollable targets: scrollable components and scrollable regions. 
% And these two targets require different identification strategies.

Scrollable components are explicit GUI elements that support swipe interactions, such as sliders and progress bars.
These components can be directly identified via heuristic parsing of XML layouts.
If a node's \texttt{scrollable} attribute is set to \texttt{true} or its \texttt{class} includes \texttt{RecyclerView}, \texttt{ScrollView} or \texttt{SeekBar}, its corresponding component is regarded as scrollable.
% Subsequently, it localizes the component using the node's \texttt{bounds} attribute.

Scrollable regions, in contrast, refer to layout-level areas that support exploratory swipes, such as content feeds, lists, and icon grids.
Unlike scrollable components, these regions often do not correspond to explicit nodes in GUI hierarchy files.
Instead, they are defined by their visual layout and semantic function.
We therefore leverage a VLM to directly infer scrollable regions from GUI screenshots.
Specifically, we use \texttt{Qwen3-VL-4B-Instruct}~\cite{qwen3}, an advanced multimodal model with strong GUI understanding capability, to reason over visual appearance.
% This enables \texttt{SwipeGen} to identify scrollable regions in a flexible and app-agnostic manner.
The prompt design is shown in Figure~\ref{fig:3_region_prompt}.

\begin{figure}[t]
    \begin{promptbox}
    \textbf{Task:} \\
    You are given a screenshot of a mobile app UI.
    Your task is to identify all scrollable regions on the screen, \ie areas that support horizontal or vertical swipes, such as content feeds, lists, carousels, or grid layouts.    
    
    \textbf{Output:} \\
    For each region, provide the following information: \\
    1) Type: a brief description of the region (\eg list, feed, carousel).
    2) Direction: if scrollable, the supported swipe direction ("horizontal", "vertical", or "both").
    3) Bbox: bounding box coordinates [x1, y1, x2, y2], where x and y range from 0 to 1000.
    4) Description: a short description of the intended interaction.
    Output the result in JSON format. Only output valid JSON. Do not include any additional text.

    \textbf{Example:}\begin{verbatim} 
[{
    "Type": "content feed",
    "Direction": "vertical",
    "Bbox": [0, 0, 1000, 1000],
    "Description": "Swipe up or down to view 
    more content in the feed"
}, ... ]
\end{verbatim}
    
    \end{promptbox}
\vspace{-9pt}
\caption{The prompt for identifying scrollable regions.}
\label{fig:3_region_prompt}
\end{figure}

\subsection{Candidate Swipe Generation}
Given the identified scrollable targets, \texttt{SwipeGen} generates candidates according to the target type.

\subsubsection{Scrollable Components.}
For scrollable components, we generate candidates as follows.
Let $b = (x_1, y_1, x_2, y_2)$ denote the component bounding box.
We set the swipe start position to the box center,
\begin{equation}
s = (s_1, s_2) = \left(\tfrac{x_1+x_2}{2}, \tfrac{y_1+y_2}{2}\right)
\end{equation}
The swipe orientation is determined by the component aspect ratio: vertical swipes are considered if $(y_2-y_1) > (x_2-x_1)$, and horizontal swipes otherwise.
A swipe direction \texttt{dir} is then randomly sampled from the valid orientations (up and down for vertical swipes, left and right for horizontal swipes).
The swipe distance is randomly sampled as a fraction of the screen size.

Specifically, let $\alpha \in (0, 1]$ denote a random scaling factor.
For a horizontal(vertical) swipe, the distance is defined as
\begin{equation}
d = \alpha \cdot L,
\end{equation}
where $L$ is the screen width or height.
Given the sampled direction, the end position is obtained by translating the start position $s$ along the swipe direction by distance $d$, while ensuring the end position remains within the screen boundary.
For a horizontal swipe, the end position is computed as
\begin{equation}
e =
\begin{cases}
(\min(s_1 + d,\, W),\, s_2), & \text{if direction is right}, \\
(\max(s_1 - d,\, 0),\, s_2), & \text{if direction is left},
\end{cases}
\end{equation}

and for a vertical swipe,
\begin{equation}
e =
\begin{cases}
(s_1,\, \min(s_2 + d,\, H)), & \text{if direction is down}, \\
(s_1,\, \max(s_2 - d,\, 0)), & \text{if direction is up},
\end{cases}
\end{equation}
As discussed in Section~\ref{ssec:unified_swipe}, duration plays a limited role for such swipes. Therefore, we fix the duration $t$ to a default value of 300ms.
Overall, each component yields 2 candidate swipes with different swipe directions, represented as $(s, e, \text{dir})$.

\subsubsection{Scrollable Regions.}
Similarly, for each scrollable region with bounding box $b$, we first compute its center position $c$.
We identify the dominant axis of the region by comparing its height and width.
If $(y_2-y_1) > (x_2-x_1)$, the region is treated as vertically scrollable. Otherwise, it is treated as horizontally scrollable.
Then, we offset the center $c$ along the dominant axis to generate candidate start positions close to the region's boundary.
Specifically, let $\alpha \in [0.2, 0.5)$ denote a random offset ratio.
For a horizontally scrollable region, the candidate start positions are
\begin{equation}
s =
\begin{cases}
(c_1 + \alpha (x_2-x_1),\, c_2), \\
(c_1 - \alpha (x_2-x_1),\, c_2),
\end{cases}
\end{equation}
and for a vertically scrollable region,
\begin{equation}
s =
\begin{cases}
(c_1,\, c_2 + \alpha (y_2-y_1)), \\
(c_1,\, c_2 - \alpha (y_2-y_1)).
\end{cases}
\end{equation}
For each start position $s$, the swipe direction is set opposite to the offset direction, reflecting natural scrolling behavior (\eg offsetting right corresponds to a left swipe).
The end position is then obtained by extending the swipe along this direction until reaching the region boundary.
For horizontal swipes,
\begin{equation}
e =
\begin{cases}
(x_1,\, s_2), & \text{if direction is left}, \\
(x_2,\, s_2), & \text{if direction is right},
\end{cases}
\end{equation}
and for vertical swipes,
\begin{equation}
e =
\begin{cases}
(s_1,\, y_1), & \text{if direction is up}, \\
(s_1,\, y_2), & \text{if direction is down}.
\end{cases}
\end{equation}
Moreover, the execution outcome of swipes over a region is sensitive to swipe speed.
In real-world systems, user control over swipe gestures is inherently coarse-grained (\eg fast vs. slow swipes), rather than continuous.
This is also reflected in OS-level gesture recognizers, which typically rely on velocity thresholds to determine scrolling behavior.
Accordingly, we divide swipe duration $t$ into two categories: a fast swipe (150ms) and a slow swipe (500ms).
Overall, each region yields $2 \times 2 = 4$ candidate swipes with different start points and duration, represented as $(s, e, \text{dir}, t)$.

\subsection{Swipe Validity Verification}
\label{ssec:swipe_validity}
However, not every candidate swipe leads to a GUI response.
To ensure data quality, \texttt{SwipeGen} adopts an execute-and-verify procedure for swipe selection.
For each scrollable target, candidate swipes are executed sequentially and verified one by one, and the first swipe that induces a perceptible GUI change is retained as a valid sample.

To achieve this, we compare the visual changes between the screenshots before and after execution.
Specifically, we convert the two screenshots to grayscale, and restrict the comparison within the target area (\ie the bounding box of the scrollable component or region).
We then compute the pixel-wise absolute difference and measure the ratio of pixels whose intensity change exceeds a fixed threshold $\delta = 0.02$.
If the swipe is considered effective and retained, \texttt{SwipeGen} moves on to the next GUI. Otherwise, the current candidate is discarded, and \texttt{SwipeGen} executes the next swipe.

Moreover, to understand the contribution of this module, 
we conduct an ablation study by removing this module while keeping the remaining pipeline unchanged.
The results are shown in table~\ref{tab:3_ablation}.

\begin{table}[!t]
    \centering
    \small
    \setlength{\tabcolsep}{2.4pt}
    \renewcommand{\arraystretch}{0.88}
    \begin{tabular}{lcccc}
    \toprule
    \textbf{Variant} & \textbf{Spotify} & \textbf{DeepSeek} & \textbf{Twitter/X} & \textbf{Overall} \\
    \midrule
    \texttt{w/o Verification} 
    & 74 
    & 14
    & 87
    & 175 \\
    \texttt{SwipeGen} 
    & 54 ($\downarrow$ 27\%) 
    & 4 ($\downarrow$ 71\%) 
    & 70 ($\downarrow$ 20\%) 
    & 128 ($\downarrow$ 27\%) \\
    \bottomrule
    \end{tabular}
    \caption{Ablation of \texttt{SwipeGen}.}
    \label{tab:3_ablation}
    \vspace{-6pt}
\end{table}

\begin{figure}[t]
    \begin{promptbox}
    \textbf{Task:} \\
    You are given two screenshots of a mobile app UI, the screen before a swipe and the screen after it. 
    You are also given the executed swipe with its parameters.
    Your task is to generate a concise, step-level natural language command written in imperative form that accurately describes the performed swipe.
    
    \textbf{Output:} \\
    The generated command should:
    1. Describe a single interaction step.
    2. Based on the visual change between the two screenshots.
    3. Reflect the intent of the swipe (\eg revealing more content, scrolling a list).
    Output the result in JSON format, include the command itself ("Command") and the reason why you describe it as so ("Reason").

    \textbf{Example:}\begin{verbatim} 
{
    "Command": "Swipe up on the main feed to view 
    more content.",
    "Reason": ...
}
\end{verbatim}
    \end{promptbox}
\vspace{-9pt}
\caption{The prompt for generating swipe descriptions.}
\label{fig:3_generate_prompt}
\end{figure}

\subsection{Swipe Description Generation}
Finally, following the reverse-synthesis practice~\cite{xie2025guiexplorer, kweon2026ghostui}, for each validated swipe, \texttt{SwipeGen} generates a corresponding step-level natural language description.
Specifically, we prompt a VLM with the GUI screenshots before and after the swipe, together with the executed swipe parameters, and ask it to describe the performed interaction in natural language.
The prompt and example outputs are provided in Figure~\ref{fig:3_generate_prompt}.

\section{SwipeBench}
\label{sec:benchmark}

In order to further advance research on swipe-aware GUI agents, we introduce \texttt{SwipeBench}.
It is the first benchmark specifically designed to evaluate swipe execution in GUI agents.
\texttt{SwipeBench} focuses on assessing the performance of GUI agents in out-of-domain (OOD) scenarios, thereby minimizing potential data leakage from VLM pretraining.
Specifically, all the selected apps in \texttt{SwipeBench} are newly released, ensuring no overlap with commonly used or previously benchmarked mobile apps, which are often included in large-scale vision-language pretraining corpora.

\texttt{SwipeGen} initially generated 421 swipe interactions from 16 distinct mobile apps, exploring key screens reachable from the home page through main navigation tabs.
After manual filtering to remove instances containing sensitive user information (\eg personal profiles) and swipes without page responses, \texttt{SwipeBench} comprises 382 valid interactions.
Each swipe data is multimodal, including images before and after the interaction, a natural language description of the interaction, the type of interaction, and any necessary parameters for the interaction.
The distribution of these swipes across different applications can be found at our GitHub repository.
\section{Evaluation}
\label{sec:evaluation}

This section evaluates the data generation capability of \texttt{SwipeGen} and the effectiveness of its generated data using \texttt{GUISwiper}.

\begin{table*}[!t]
    \centering
    \begin{tabular}{lccc}
    \toprule
    \textbf{Module} & \textbf{Avg. Time} & \textbf{Prompt Token} & \textbf{Completion Token} \\
    \midrule
    GUI Exploration & 38.82 s / page & -- & -- \\
    Scrollable Target Identification & 13.90 s / page & 2860 / page & 67 / page \\
    % Candidate Swipe Generation & -- & -- & -- \\
    Swipe Validity Verification & 4.34 s / swipe & -- & -- \\
    Swipe Description Generation & 13.47 s / swipe & 5668 / swipe & 57 / swipe \\
    \midrule
    \textbf{Total} & 120.46 s / swipe & 9374 / swipe & 144 / swipe \\
    \bottomrule
    \end{tabular}
    \caption{Runtime and token cost of \texttt{SwipeGen}.}
    \label{tab:5_overhead}
    \vspace{-6pt}
\end{table*}

\subsection{GUISwiper}
We fine-tuned \texttt{GUISwiper}, a GUI-specific VLM capable of performing multiple types of interactions with GUIs, particularly swipes.
To enhance generalization while keeping training costs low, we employed the advanced RLVR method~\cite{lu2025uir1, luo2025guir1, liu2025infiguig1, chen2025uiins, tang2025guig2}.
Additionally, to assess the quality of the data generated by \texttt{SwipeGen}, we selected one of the most widely used base models, Qwen2.5-VL-3B-Instruct~\cite{Qwen2.5-VL}.

\subsubsection{Training Dataset}
The training dataset consists of 927 swipe interactions and 579 tap interactions, collected from 43 popular applications.
% These apps cover various domains such as communication, productivity, entertainment, and system utilities.
The apps were selected based on their ranking within Google Play categories, ensuring that the most popular app in each category was included.
The swipe data was automatically generated using \texttt{SwipeGen}, while the tap data was collected during the GUI exploration phase and later annotated by a VLM.

\subsubsection{Runtime Analysis}
We additionally analyze the computational cost of \texttt{SwipeGen}.
Table~\ref{tab:5_overhead} reports the runtime and token consumption of each module when generating the GUISwiper training set.

\subsubsection{Reward Design}
We design the overall reward as a combination of three components: a format reward, an action type reward, and an action accuracy reward.
The final reward is linearly normalized to $[-1, 1]$ for stable training.

% \textit{Format Reward.}
We adopt a widely used format reward~\cite{lu2025uir1, zhang2025btlui} to enforce structured outputs.
The model is required to produce reasoning enclosed in \texttt{<think>} tags followed by a final answer in a predefined JSON format.
Correctly formatted outputs receive a reward of $+1$, while malformed outputs receive $-1$.
% We assign a relatively high weight to this reward since correct formatting is a prerequisite for executable GUI interactions.

% \textit{Action Type Reward.}
To prevent overfitting to swipes and preserve general GUI interaction capabilities, we include an action type reward.
If the predicted action type (\eg \texttt{swipe}, \texttt{click}, \texttt{input}) matches the ground truth, the model receives $+0.8$; otherwise, $-0.8$.

% \textit{Action Accuracy Reward.} 
For non-swipe actions, we follow the same accuracy reward design as UI-R1~\cite{lu2025uir1}.
As for swipes, we define an accuracy reward $R_{acc} \in [0, 1]$ as the sum of four sub-rewards:
\begin{equation}
    R_{acc} = R_{\text{start}} + R_{\text{end}} + R_{\text{dir}} + R_{\text{dur}}.
\end{equation}

(1) If the Euclidean distance between the predicted and ground-truth start positions is within 220 pixels, a reward of $0.45$ is given; otherwise, $0$.
Following prior work~\cite{liu2025infiguig1}, we normalize all the image resolutions to $(0, 0, 1000, 1000)$, and set a 220-pixel tolerance for allowing reasonable localization errors.
Additionally, for region-level swipes, a constraint is enforced: if the predicted start position lies outside the target bounding box, the start-position reward is set to $0$ regardless of distance.
% Let $\hat{s}$ and $s$ denote the predicted and ground-truth start positions, we define:
% \begin{equation}
%     R_{\text{start}} =
%     \begin{cases}
%     0.45, & \|\hat{s} - s\|_2 \leq 220 \text{ and } \hat{s} \in B \\
%     0, & \text{otherwise}.
%     \end{cases}
% \end{equation}
% where $B$ denotes a bounding box:
% \begin{equation}
%     B = \begin{cases}
%         (x_1,y_1,x_2,y_2), & \text{for scrollable regions} \\
%         (0,0,W,H), & \text{for scrollable components}
%     \end{cases}
% \end{equation}

% Following prior work~\cite{liu2025infiguig1}, .

(2) If the predicted end position is within 220 pixels of the ground truth, a reward of $0.10$ is given.
This term is assigned a lower weight since end positions are less important for scrollable regions.
% Let $\hat{e}$ and $e$ denote the predicted and ground-truth end positions:
% \begin{equation}
%     R_{\text{end}} =\begin{cases}
%     0.10, & \|\hat{e} - e\|_2 \leq 220 \\
%     0, & \text{otherwise}.
%     \end{cases}
% \end{equation}

(3) If the predicted swipe direction (\eg \texttt{up}) matches the ground truth, a reward of $0.35$ is given.
% Let $\hat{\text{dir}}$ and $\text{dir}$ denote the predicted and ground-truth swipe directions:
% \begin{equation}
%     R_{\text{dir}} =
%     \begin{cases}
%     0.35, & \hat{\text{dir}} = \text{dir} \\
%     0, & \text{otherwise}.
%     \end{cases}
% \end{equation}

(4) For component-level swipes, the reward of $0.1$ is always granted.
For region-level swipes, we discretize duration into fast (150\,ms) and slow (500\,ms).
Predicted durations are mapped to these categories using a midpoint threshold, and a reward of $0.10$ is given if the predicted category matches the ground truth.

Notably, all accuracy rewards are non-negative: incorrect predictions do not incur penalties, which is crucial for stable training.

% \subsubsection{Implementations}
% We fine-tune Qwen2.5-VL-3B-Instruct using RLVR-based training with the Adam optimizer in BFloat16 precision. 
% To stabilize training while preserving the visual details required for GUI interaction, we set the maximum image resolution to 1,048,576 pixels and use a relatively small learning rate decayed from $9.9\times10^{-7}$ to $5.0\times10^{-7}$. 
% Training is conducted for 8 epochs with a per-device batch size of 1 and gradient accumulation over 2 steps. 
% Following prior RLVR-based GUI agents, we sample 8 generations for each training instance to improve exploration and reward optimization.

\begin{table}[t]
\centering
\small
\begin{tabular}{cccc}
\toprule
\textbf{Model} & \textbf{Model Size} & \textbf{Success Rate (\%)} \\
\midrule
Qwen2.5-VL~\cite{Qwen2.5-VL} & 3B & 24.87 \\
Qwen3-VL~\cite{qwen3} & 2B & 39.79 \\
UGround~\cite{gou2025uground} & 2B & 18.85 \\
UI-R1~\cite{lu2025uir1} & 3B & 30.41 \\
GUI-Owl~\cite{ye2025mobileagentv3} & 2B & 30.37 \\
UI-TARS~\cite{qin2025uitars} & 2B & \underline{52.09} \\
MAI-UI~\cite{zhou2025maiui} & 2B & 48.95 \\
\midrule
\textbf{GUISwiper} & 3B & \textbf{61.25} \\
\bottomrule
\end{tabular}
\caption{Swipe execution success rate on \texttt{SwipeBench}.}
\label{table:5_RQ1}
\vspace{-6pt}
\end{table}

\begin{table*}[t]
\centering
\small
\begin{tabular}{ccccc}
\toprule
\textbf{Model} & \textbf{Model Size} & \textbf{Text Accuracy (\%)} & \textbf{Icon/Widget Accuracy (\%)} & \textbf{Avg. (\%)} \\
\midrule
Qwen2.5-VL~\cite{Qwen2.5-VL} & 3B & 90.50 & 61.10 & 75.80 \\
% Qwen3-VL~\cite{qwen3} & 2B &  &  &  \\
UGround~\cite{gou2025uground} & 2B & 82.80 & 60.30 & 71.60 \\
UI-R1~\cite{lu2025uir1} & 3B & \textbf{95.60} & \textbf{84.70} & \textbf{90.15} \\
% GUI-Owl~\cite{ye2025mobileagentv3} & 2B &  &  &  \\
UI-TARS~\cite{qin2025uitars} & 2B & 93.00 & 75.50 & 84.25 \\
% MAI-UI~\cite{zhou2025maiui} & 2B &  &  &  \\
\midrule
\textbf{GUISwiper} & 3B & \underline{91.20} & \underline{80.32} & \underline{85.83} \\
\bottomrule
\end{tabular}
\caption{Grounding accuracy on ScreenSpot.}
\label{table:5_RQ2}
\vspace{-6pt}
\end{table*}

\begin{table}[t]
\centering
\small
\begin{tabular}{ccc}
\toprule
\textbf{Model} & \textbf{Model Size} & \textbf{Success Rate (\%)} \\
\midrule
\multicolumn{3}{c}{\textit{Comparable-scale Models}} \\
\midrule
Qwen3-VL & 2B & 36.40 \\
UGround & 2B & 32.80 \\
MAI-UI & 2B & 49.10 \\
GUI-Owl & 2B & 73.30 \\
\midrule
\multicolumn{3}{c}{\textit{Large-scale Models}} \\
\midrule
UI-TARS & 7B & 33.00 \\
MAI-UI & 8B & 70.70 \\
Qwen3-VL & 8B & 47.60 \\
Qwen3-VL & 32B & 57.30 \\
Qwen2.5-VL & 32B & 22.00 \\
Qwen2.5-VL & 72B & 35.00 \\
UI-TARS & 72B & 46.60 \\
\midrule
\textbf{GUISwiper} & 3B & 40.52 \\
\bottomrule
\end{tabular}
\caption{End-to-end task success rate on AndroidWorld.}
\label{table:5_RQ3}
\vspace{-6pt}
\end{table}

\subsection{Experimental Settings}
\subsubsection{Benchmarks}
We evaluate GUI agents on three benchmarks: \texttt{SwipeBench}, ScreenSpot~\cite{cheng2024seeclick} and  AndroidWorld~\cite{rawles2025androidworld}.

ScreenSpot is a widely used GUI grounding evaluation benchmark. We evaluate \texttt{GUISwiper} on its mobile subset.

AndroidWorld is an online benchmark consisting of 116 multi-step GUI automation tasks. Unlike offline grounding benchmarks, AndroidWorld requires agents to interact with dynamic environments through sequential decision-making and action execution. 
    % We evaluate \texttt{GUISwiper} on AndroidWorld to examine whether improved swipe capability can translate into better downstream task completion in realistic online settings.

\subsubsection{Baselines}
We compare \texttt{GUISwiper} against multiple state-of-the-art (SOTA) GUI-specific VLMs.  
\begin{itemize}
    \item Qwen2.5-VL-3B-Instruct~\cite{Qwen2.5-VL}, the foundation model.
    \item Qwen3-VL-2B-Instruct~\cite{qwen3}, the latest one in the Qwen series. It is specifically trained on GUI data.
    \item UGround~\cite{gou2025uground} is trained via supervised fine-tuning (SFT) on a dataset comprising 10M GUI elements and over 1.3M screenshots.
    \item UI-R1~\cite{lu2025uir1} and GUI-Owl~\cite{ye2025mobileagentv3} employ RL-based training, representing recent advances in this paradigm.
    \item UI-TARS~\cite{qin2025uitars} and MAI-UI~\cite{zhou2025maiui} are industrial SOTA models trained on large-scale proprietary GUI data.
\end{itemize}

\subsection{Experimental Result and Analysis}

\subsubsection{Swipe Execution}
Table~\ref{table:5_RQ1} reports the swipe execution success rate on \texttt{SwipeBench}.
We formulate swipe execution as a binary classification task, where a swipe is considered successful only if all the parameters satisfy the same criteria used during training.

As shown in the table, directly prompting a Qwen2.5-VL yields a low success rate of 24.87\%.
In contrast, \texttt{GUISwiper} improves the success rate to 61.25\%, achieving a 2.46$\times$ relative improvement and outperforming all baselines, including UI-TARS (52.09\%) and MAI-UI (48.95\%).

% Notably, \texttt{GUISwiper} consistently outperforms all baselines, including industrial models such as UI-TARS (52.09\%) and MAI-UI (48.95\%), demonstrating that improving swipe interaction quality does not rely on large-scale data, but can be effectively achieved through high-quality synthetic supervision.

Furthermore, \texttt{GUISwiper} successfully completes 23 tasks where all baseline models fail.
These tasks mainly involve long-range scrolling, feed navigation, and navigation exploration, highlighting the importance of reliable swipe execution for complex GUI interactions.
% This supports our claim that human-like swipe modeling is critical for robust GUI interaction.

Moreover, we analyze the failure cases of \texttt{GUISwiper} to understand the remaining challenges.
Region-level errors are the dominant failure source (69\%), where the model predicts swipe locations outside valid interaction regions.
Distance estimation errors account for another 25\% of failures, especially when the instruction does not explicitly specify swipe magnitude.
In contrast, direction errors only account for 6\%, suggesting that coarse directional understanding is no longer the primary bottleneck.
% First, region-level reasoning emerges as the dominant source of error, accounting for 69\% of all failures. % 100 / 145
% In these cases, the model selects a starting point outside the valid scrollable region. 
% This indicates that successful swipe execution requires jointly reasoning about spatial constraints (\ie valid interaction regions) and user intent, rather than merely identifying visually salient elements. 
% Second, distance errors constitute the second largest category, accounting for 25\% of failures. % 36 / 145
% These errors typically occur when the instruction dose not specify the swipe distance, requiring the model to infer implicit distance based on the GUI context. 
% This suggests that while \texttt{SwipeGen} improves spatial control, modeling fine-grained magnitude remains challenging.
% Finally, only 6\% of failures are attributed to incorrect swipe directions. % 9 / 145
% This relatively low error rate indicates that coarse directional understanding (\ie up, down) is not the primary bottleneck in swipe execution.

\subsubsection{Grounding Capability}

Table~\ref{table:5_RQ2} reports grounding accuracy on ScreenSpot. 
Our primary goal is to examine whether fine-tuning on interaction data degrades general GUI grounding capability.
Some recent models (\eg Qwen3-VL, GUI-Owl, MAI-UI) do not report results on ScreenSpot, and are therefore excluded from this comparison.

As shown in the table, \texttt{GUISwiper} achieves an overall accuracy of 85.83\%, outperforming its backbone model Qwen2.5-VL (75.80\%) while remaining competitive with strong baselines such as UI-TARS (84.25\%) and UI-R1 (90.15\%). 
These results suggest that enhancing region-level swipe capabilities does not come at the cost of degraded element localization.
Overall, the results demonstrate that swipe execution and grounding capability can be jointly maintained without trade-offs.

\subsubsection{End-to-end Task Execution}
% 本质是检查最终任务是否完成，可以看到每个任务进行的操作的占比

Table~\ref{table:5_RQ3} reports end-to-end task success rates on AndroidWorld.
\texttt{GUISwiper} achieves a success rate of 40.52\%, outperforming comparable-scale models such as Qwen3-VL-2B (36.40\%) and UGround (32.80\%), as well as larger backbone models including Qwen2.5-VL-32B (22.00\%) and Qwen2.5-VL-72B (35.00\%).
These results demonstrate that improving swipe execution can translate into better downstream task completion.
Note that GUI-Owl achieves higher performance through the MobileAgent-v3 workflow with additional agentic components, while \texttt{GUISwiper} focuses only on low-level swipe interaction.

We further analyze \texttt{GUISwiper}'s execution traces and find that successful tasks often involve long-horizon off-screen navigation.
For example, in \texttt{RecipeAddMultipleRecipes}, the agent repeatedly scrolls through long recipe pages while continuously entering content into the Broccoli app, with scroll actions accounting for 41.9\% of all executed steps. 
Similar tasks include \texttt{RecipeAddSingleRecipe}, \texttt{ExpenseAddMultiple}, etc.
These cases show that reliable swipe execution is important for maintaining progress when agents repeatedly access invisible UI content.

Overall, these results indicate that swipe execution is an important bottleneck for end-to-end GUI automation.
\section{Discussion}
\label{sec:discussion}

\subsection{Limitations}

While \texttt{SwipeGen} improves GUI agents' swipe execution capability, we acknowledge several limitations. 
First, our definition of human-like focuses on the task-effect level, \ie whether a swipe aligns with user intent and induces the expected GUI transition. 
It does not aim to reproduce the complete motor characteristics of human swipes, such as trajectory curvature or velocity variation.

Second, our Swipe Validity Verification module adopts lightweight pixel-level comparison to enable scalable data synthesis. 
Although we restrict the comparison area to the target scrollable component or region, dynamic content within this area may still introduce false positives.

\subsection{Future Work}

We identify several promising directions for future work. 
First, we plan to investigate more fine-grained human swipe modeling by incorporating real user interaction traces, including trajectory patterns and temporal dynamics. 
Such modeling could further narrow the gap between GUI agents and human users.

Second, we plan to scale up both synthesized training data and evaluation benchmarks to cover more diverse applications and long-horizon interaction scenarios.
\section{Conclusion}
\label{sec:conclusion}
In this work, we address a critical yet overlooked limitation in GUI agents: their inability to execute swipe interactions effectively due to prevailing component-centric interaction strategies. 

Our contributions are three-fold. 
First, we propose \texttt{SwipeGen}, the first automated pipeline for synthesizing human-like swipe interactions, enabling proper formulation of swipe data that has been lacking in existing datasets. 
Second, we introduce \texttt{SwipeBench}, the first benchmark for evaluating swipe interaction quality, comprising 382 swipes from 16 OOD mobile apps. 
Third, through \texttt{GUISwiper}, we show that VLMs fine-tuned with synthesized swipe data can substantially address the swipe execution bottleneck.
% We hope our work highlights the importance of human gesture modeling in GUI agents and could encourage future research to move toward human-like interaction execution.

\begin{acks}
This work is supported by the National Natural Science Foundation of China (Project No. 62572127). Y.Zhou is the corresponding author.
\end{acks}

\bibliographystyle{ACM-Reference-Format}
\bibliography{ref.bib}

\end{document}